\documentclass[runningheads]{llncs}
\usepackage{graphicx}
\usepackage{amsmath,amssymb} 
\usepackage{color}
\usepackage{epsfig}
\usepackage{algorithm}
\usepackage{algorithmic}
\usepackage{multirow}
\usepackage[width=122mm,left=12mm,paperwidth=146mm,height=193mm,top=12mm,paperheight=217mm]{geometry}
\begin{document}
\pagestyle{headings}
\mainmatter

\title{A Unified Multi-scale Deep Convolutional Neural Network for Fast Object Detection} 

\titlerunning{A Unified Multi-scale Deep CNN for Fast Object Detection}

\authorrunning{Zhaowei Cai, Quanfu Fan, Rogerio S. Feris, and Nuno Vasconcelos}

\author{Zhaowei Cai$^1$\and Quanfu Fan$^2$\and Rogerio S. Feris$^2$\and Nuno Vasconcelos$^1$}


\institute{$^1$SVCL, UC San Diego\\
$^2$IBM T. J. Watson Research\\
\{zwcai,nuno\}@ucsd.edu, \{qfan,rsferis\}@us.ibm.com}

\maketitle

\begin{abstract}
A unified deep neural network, denoted the {\it multi-scale CNN\/} (MS-CNN),
is proposed for fast multi-scale object detection. The MS-CNN
consists of a proposal sub-network and a detection sub-network. In the
proposal sub-network, detection is performed at multiple output layers,
so that receptive fields match objects of different scales. These
complementary scale-specific detectors are combined to produce a strong
multi-scale object detector. The unified network is learned end-to-end, by
optimizing a multi-task loss. Feature upsampling by deconvolution is also
explored, as an alternative to input upsampling, to reduce the memory and
computation costs. State-of-the-art object detection performance, at up
to 15 fps, is reported on datasets, such as KITTI and Caltech, containing a substantial number of small objects.

\end{abstract}

\section{Introduction}

\begin{figure}[!t]
\centering
\centerline{\epsfig{figure=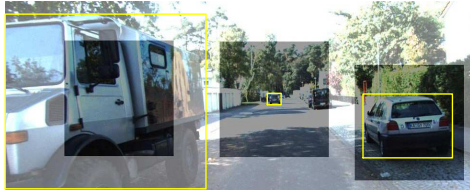,width=6.25cm,height=2.5cm}}
\caption{In natural images, objects can appear at very different
scales, as illustrated by the yellow bounding boxes.
A single receptive field, such as that of the RPN
\cite{DBLP:conf/nips/shaoqing15fasterRcnn} (shown in the shaded area),
cannot match this variability.}
\label{fig:receptive field}
\end{figure}

Classical object detectors, based on the sliding window paradigm,
search for objects at multiple scales and aspect ratios. While real-time
detectors are available for certain classes of objects, e.g. faces or
pedestrians \cite{DBLP:journals/ijcv/ViolaJ04,DBLP:journals/pami/DollarABP14},
it has proven difficult to build detectors of multiple object classes
under this paradigm.
Recently, there has been interest in detectors derived from deep
convolutional neural networks (CNNs)~\cite{DBLP:conf/cvpr/GirshickDDM14,DBLP:conf/iccv/Girshick15,DBLP:conf/nips/XiaozhiNIPS15,DBLP:conf/eccv/HeZR014,DBLP:conf/iccv/GidarisK15}.  While these have shown much greater ability to address the
multiclass problem, less progress has been made towards the detection
of objects at multiple scales. The R-CNN \cite{DBLP:conf/cvpr/GirshickDDM14}
samples object proposals at multiple scales, using a preliminary attention
stage \cite{DBLP:conf/iccv/SandeUGS11}, and then warps these proposals to
the size (e.g. 224$\times$224) supported by the CNN. This is, however,
very inefficient from a computational standpoint. The development of
an effective and computationally efficient region proposal mechanism
is still an open problem. The more recent
Faster-RCNN \cite{DBLP:conf/nips/shaoqing15fasterRcnn} addresses the issue
with a region proposal network (RPN), which enables end-to-end
training. However, the RPN generates proposals of multiple scales by sliding a
fixed set of filters over a fixed set of convolutional feature maps.
This creates an inconsistency between the sizes of objects, which are
variable, and filter receptive fields, which are fixed. As shown in
Fig. \ref{fig:receptive field}, a fixed receptive field cannot cover
the multiple scales at which objects appear in natural scenes.
This compromises detection performance, which tends to be particularly poor
for small objects, like that in the center of Fig. \ref{fig:receptive field}.
In fact, \cite{DBLP:conf/iccv/Girshick15,DBLP:conf/nips/XiaozhiNIPS15,DBLP:conf/nips/shaoqing15fasterRcnn} handle such objects by upsampling the input
image both at training and testing time. This increases the memory
and computation costs of the detector.

This work proposes a unified multi-scale deep CNN,
denoted the {\it multi-scale CNN\/} (MS-CNN), for fast object detection.
Similar to \cite{DBLP:conf/nips/shaoqing15fasterRcnn}, this
network consists of two sub-networks: an object proposal network and an
accurate detection network. Both of them are learned end-to-end and share
computations. However, to ease the inconsistency between the sizes of
objects and receptive fields, object detection is performed with multiple
output layers, each focusing on objects within certain scale ranges
(see Fig. \ref{fig:framework}). The intuition is that lower network layers,
such as ``conv-3,'' have smaller receptive fields, better matched
to detect small objects. Conversely, higher layers, such as ``conv-5,''
are best suited for the detection of large objects. The complimentary
detectors at different output layers are combined to form a strong
multi-scale detector. This is shown to produce accurate object
proposals on detection benchmarks with large variation
of scale, such as KITTI \cite{DBLP:conf/cvpr/GeigerLU12},
achieving a recall of over 95\% for only 100 proposals.

A second contribution of this work is the use of feature upsampling as an
alternative to input upsampling. This is achieved by introducing a
deconvolutional layer that increases the resolution of feature maps
(see Fig. \ref{fig:frcnn}), enabling small objects to produce larger
regions of strong response. This is shown to  reduce memory and computation
costs. While deconvolution has been explored for segmentation
\cite{DBLP:conf/cvpr/LongSD15} and edge
detection \cite{DBLP:conf/iccv/XieT15}, it is, as far as we know, for
the first time used to speed up and improve detection. When combined
with efficient context encoding and hard negative mining, it results
in a detector that advances the state-of-the-art detection on
the KITTI \cite{DBLP:conf/cvpr/GeigerLU12} and
Caltech \cite{DBLP:journals/pami/DollarWSP12} benchmarks.
Without image upsampling, the MS-CNN achieves speeds of 10 fps on KITTI
(1250$\times$375) and 15 fps on Caltech (640$\times$480) images.


\section{Related Work}

One of the earliest methods to achieve real-time detection with  high
accuracy was the cascaded detector of \cite{DBLP:journals/ijcv/ViolaJ04}.
This architecture has been widely used to implement sliding window
detectors for faces \cite{DBLP:journals/ijcv/ViolaJ04,DBLP:conf/cvpr/BourdevB05}, pedestrians \cite{DBLP:journals/pami/DollarABP14,DBLP:conf/iccv/CaiSV15} and cars \cite{DBLP:journals/tits/Ohn-BarT15}.
Two main streams of research have been pursued to improve its speed:
fast feature extraction \cite{DBLP:journals/ijcv/ViolaJ04,DBLP:journals/pami/DollarABP14} and cascade learning
\cite{DBLP:conf/cvpr/BourdevB05,DBLP:journals/jmlr/SaberianV14,DBLP:conf/iccv/CaiSV15}. In \cite{DBLP:journals/ijcv/ViolaJ04}, a set
of efficient Haar features was proposed with recourse to integral images.
The aggregate feature channels (ACF) of~\cite{DBLP:journals/pami/DollarABP14}
made it possible to compute HOG features at about 100 fps. On the learning
front,~\cite{DBLP:conf/cvpr/BourdevB05} proposed the soft-cascade, a
method to transform a classifier learned with boosting  into a cascade with
certain guarantees in terms of false positive and detection rate.
\cite{DBLP:journals/jmlr/SaberianV14} introduced a Lagrangian formulation
to learn cascades that achieve the optimal trade-off between accuracy and
computational complexity. \cite{DBLP:conf/iccv/CaiSV15} extended this
formulation for cascades of highly heterogeneous features, ranging from ACF
set to deep CNNs, with widely different complexity. The main current
limitation of detector cascades is the difficulty of implementing multiclass
detectors under this architecture.

In an attempt to leverage the success of deep neural networks for object
classification, \cite{DBLP:conf/cvpr/GirshickDDM14} proposed the R-CNN
detector. This combines an object proposal mechanism
\cite{DBLP:conf/iccv/SandeUGS11} and a CNN classifier
\cite{DBLP:conf/nips/KrizhevskySH12}. While the R-CNN surpassed previous
detectors \cite{DBLP:journals/pami/FelzenszwalbGMR10,DBLP:conf/iccv/WangYZL13}
by a large margin, its speed is limited by the
need for object proposal generation and repeated CNN evaluation.
\cite{DBLP:conf/eccv/HeZR014} has shown that this could be ameliorated
with recourse to spatial pyramid pooling (SPP), which allows the computation
of CNN features once per image, increasing
the detection speed by an order of magnitude. Building on SPP,
the Fast-RCNN \cite{DBLP:conf/iccv/Girshick15} introduced
the ideas of back-propagation through the ROI pooling layer and multi-task
learning of a classifier and a bounding box regressor. However, it still
depends on bottom-up proposal generation. More recently, the
Faster-RCNN \cite{DBLP:conf/nips/shaoqing15fasterRcnn} has addressed
the generation of object proposals and classifier within a single
neural network, leading to a significant speedup for proposal detection.
Another interesting work is YOLO \cite{DBLP:conf/cvpr/RedmonDGF16}, which
outputs object detections within a 7$\times$7 grid. This network runs
at $\sim$40 fps, but with some compromise of detection accuracy.

For object recognition, it has been shown beneficial to
combine multiple losses, defined on intermediate layers of a single
network \cite{DBLP:conf/cvpr/SzegedyLJSRAEVR15,DBLP:conf/aistats/LeeXGZT15,DBLP:conf/cvpr/LongSD15,DBLP:conf/iccv/XieT15}.
GoogLeNet \cite{DBLP:conf/cvpr/SzegedyLJSRAEVR15} proposed the use of three
weighted classification losses, applied at layers of intermediate heights,
showing that this type of regularization is useful for very deep models.
The deeply supervised network architecture of
\cite{DBLP:conf/aistats/LeeXGZT15} extended this idea to a larger number of
layers. The fact that higher layers convey more semantic information
motivated \cite{DBLP:conf/cvpr/LongSD15} to combine features from
intermediate layers, leading to more accurate semantic segmentation.
A similar idea was shown useful for edge detection
in \cite{DBLP:conf/iccv/XieT15}. Similar to
\cite{DBLP:conf/cvpr/SzegedyLJSRAEVR15,DBLP:conf/aistats/LeeXGZT15,DBLP:conf/cvpr/LongSD15,DBLP:conf/iccv/XieT15},
the proposed MS-CNN is learned with losses that account for
intermediate layer outputs. However, the aim is not to simply regularize
the learning, as in \cite{DBLP:conf/cvpr/SzegedyLJSRAEVR15,DBLP:conf/aistats/LeeXGZT15}, or provide detailed information for higher
outputs, as in \cite{DBLP:conf/cvpr/LongSD15,DBLP:conf/iccv/XieT15}.
Instead, the goal is to produce a strong individual object detector
at each intermediate output layer.

\begin{figure}[!t]
\centering
\centerline{\epsfig{figure=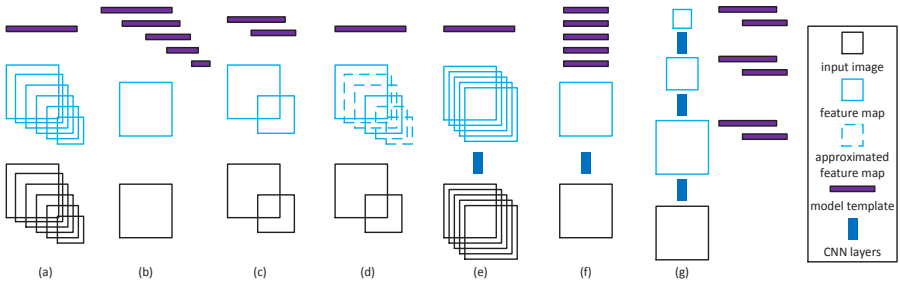,width=12cm,height=3.7cm}}
\caption{Different strategies for multi-scale detection. The length of model template represents the template size.}
\label{fig:multi-scale}
\end{figure}

\section{Multi-scale Object Proposal Network}
\label{sec:proposal network}

In this section, we introduce the proposed network for the
generation of object proposals.

\subsection{Multi-scale Detection}

The coverage of many object scales is a critical problem for object
detection. Since a detector is basically a dot-product between a learned
template and an image region, the template has to be matched to the
spatial support of the object to recognize. There are two main strategies
to achieve this goal. The first is to learn a single classifier and
rescale the image multiple times, so that the classifier can match all
possible object sizes. As illustrated in Fig. \ref{fig:multi-scale} (a),
this strategy requires feature computation at multiple image scales.
While it usually produces the most accurate detection, it tends to
be very costly. An alternative approach is to apply multiple classifiers
to a single input image. This strategy, illustrated in
Fig. \ref{fig:multi-scale} (b), avoids the repeated computation of feature
maps and tends to be efficient. However, it requires an individual classifier
for each object scale and usually fails to produce good detectors.
Several approaches have been proposed to achieve a good trade-off between
accuracy and complexity. For example, the strategy of
Fig. \ref{fig:multi-scale} (c) is to rescale the input a few times and learn
a small number of model templates \cite{DBLP:conf/cvpr/BenensonMTG12}. Another
possibility is the feature approximation
of \cite{DBLP:journals/pami/DollarABP14}. As shown in
Fig. \ref{fig:multi-scale} (d), this consists of rescaling the input a small
number of times and interpolating the missing feature maps. This
has been shown to achieve considerable speed-ups for a very modest loss of
classification accuracy \cite{DBLP:journals/pami/DollarABP14}.

The implementation of multi-scale strategies on CNN-based detectors is
slightly different from those discussed above, due to the complexity of CNN
features. As shown in Fig. \ref{fig:multi-scale} (e), the R-CNN of
\cite{DBLP:conf/cvpr/GirshickDDM14} simply warps object proposal patches to
the natural scale of the CNN. This is somewhat similar to Fig.
\ref{fig:multi-scale} (a), but features are computed for patches rather
than the entire image. The multi-scale mechanism of the
RPN \cite{DBLP:conf/nips/shaoqing15fasterRcnn}, shown in
Fig. \ref{fig:multi-scale} (f), is similar to that of
Fig. \ref{fig:multi-scale} (b). However, multiple sets of templates of the
same size are applied to all feature maps. This can lead to a
severe scale inconsistency for template matching. As shown
in Fig. \ref{fig:receptive field}, the single scale of the feature maps,
dictated by the (228$\times$228) receptive field of the CNN, can be
severely mismatched to small (e.g. 32$\times$32) or large
(e.g. 640$\times$640) objects. This compromises object detection
performance.

Inspired by previous evidence  on the benefits of the strategy of
Fig. \ref{fig:multi-scale} (c) over that of
Fig. \ref{fig:multi-scale} (b), we propose a new
multi-scale strategy, shown in Fig. \ref{fig:multi-scale} (g).
This can be seen as the deep CNN  extension of Fig. \ref{fig:multi-scale} (c),
but only uses a single scale of input. It differs from both
Fig. \ref{fig:multi-scale} (e) and (f) in that it exploits feature maps of
several resolutions to detect objects at different scales. This is
accomplished by the application of a set of templates at intermediate
network layers. This results in a set of variable receptive field sizes,
which can cover a large range of object sizes.

\begin{figure}[!t]
\centering
\centerline{\epsfig{figure=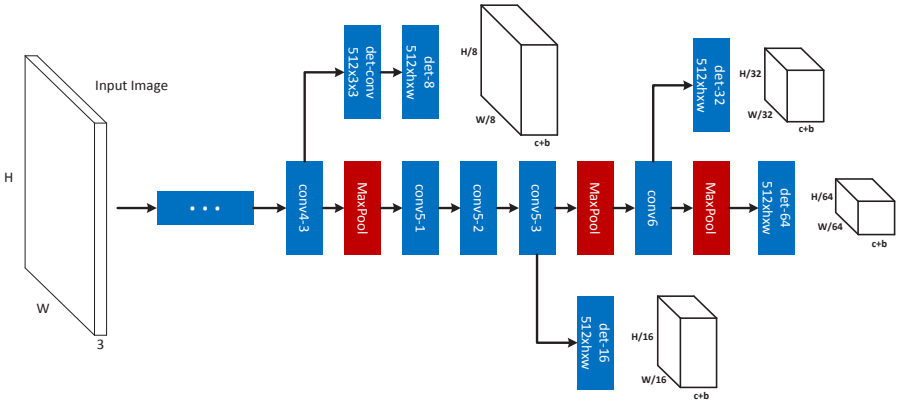,width=12cm,height=5.25cm}}
\caption{Proposal sub-network of the MS-CNN. The bold cubes are the output tensors of the network. $h\times{w}$ is the filter size, $c$ the number of classes, and $b$ the number of bounding box coordinates.}
\label{fig:framework}
\end{figure}

\subsection{Architecture}
\label{subsec:framework}

The detailed architecture of the MS-CNN proposal network is shown in
Fig. \ref{fig:framework}. The network detects objects through several
detection branches. The results by all detection branches are simply declared as the final proposal detections. The network has a standard CNN trunk, depicted in the center of
the figure, and a set of output branches, which emanate from different
layers of the trunk. These branches consist of a single detection
layer. Note that a buffer convolutional layer is introduced on the
branch that emanates after layer ``conv4-3''. Since this branch is close to
the lower layers of the trunk network, it affects their gradients more than
the other detection branches. This can lead to some instability during
learning. The buffer convolution prevents the gradients of the
detection branch from being back-propagated directly to the trunk layers.

During training, the parameters $\textbf{W}$ of the multi-scale proposal
network are learned from a set of training samples $S=\{(X_i,Y_i)\}_{i=1}^{N}$,
where $X_i$ is a training image patch, and $Y_i=(y_i,b_i)$ the combination of
its class label $y_i\in\{0,1,2,\cdots,K\}$ and bounding box
coordinates $b_i=(b_i^x,b_i^y,b_i^w,b_i^h)$. This is achieved with a
multi-task loss
\begin{equation}
\mathcal{L}(\textbf{W})
=\sum_{m=1}^{M}\sum_{i\in{S^m}}\alpha_{m}l^{m}(X_i,Y_i|\textbf{W}),
\label{equ:multi-task}
\end{equation}
where  $M$ is the number of detection branches,  $\alpha_m$ the weight of
loss $l^m$, and $S=\{S^1,S^2,\cdots,S^M\}$, where $S^m$ contains
the examples of scale $m$. Note that only a subset $S^m$ of the training
samples, selected by scale, contributes to the loss of detection layer $m$.
Inspired by the success of joint learning of classification and bounding box
regression
\cite{DBLP:conf/iccv/Girshick15,DBLP:conf/nips/shaoqing15fasterRcnn}, the
loss of each detection layer combines these two objectives
\begin{equation}
\label{equ:cls and bb}
l(X,Y|\textbf{W})=L_{cls}(p(X),y)+\lambda[y\geq{1}]L_{loc}(b,\hat{b}),
\end{equation}
where $p(X)=(p_0(X),\cdots,p_K(X))$ is the probability distribution over
classes, $\lambda$ a trade-off coefficient, $L_{cls}(p(X),y)=-\log p_{y}(X)$ the
cross-entropy loss, $\hat{b}=(\hat{b}_x,\hat{b}_y,\hat{b}_w,\hat{b}_h)$ the
regressed bounding box, and
\begin{equation}
L_{loc}(b,\hat{b})=\frac{1}{4}\sum_{j\in\{x,y,w,h\}}smooth_{L_{1}}(b_j,\hat{b}_j),
\end{equation}
the smoothed bounding box regression loss of \cite{DBLP:conf/iccv/Girshick15}.
The bounding box loss is only used for positive samples and the
optimal parameters $\textbf{W}^*=\arg\min_{\textbf{W}}\mathcal{L}(\textbf{W})$
are learned by stochastic gradient descent.

\subsection{Sampling}
This section describes the assembly of training samples
$S^m=\{S_+^m,S_-^m\}$ for each detection layer $m$. In what follows,
the superscript $m$ is dropped for notional simplicity. An anchor is centered at the sliding window on layer $m$ associated with width and height corresponding to filter size. More details can be found in Table \ref{tab:parameters}. A sample $X$ of anchor bounding box $b$ is labeled as positive if $o^*\geq{0.5}$, where
\begin{equation}
o^*=\max_{i\in{S_{gt}}}IoU(b,b_i).
\end{equation}
$S_{gt}$ is the ground truth and $IoU$ the intersection over union between
two bounding boxes. In this case, $Y=(y_{i^*},b_{i^*})$, where $i^*=\arg\max_{i\in{S_{gt}}}IoU(b,b_i)$ and $(X, Y)$ are added to
the positive set $S_+$. All the positive samples in $S_+=\{(X_i,Y_i)|y_i\geq{1}\}$ contribute to the loss. Samples such that $o^*<0.2$ are assigned to a preliminary negative training pool, and the remaining samples discarded. For a natural image, the distribution of objects and non-objects is heavily asymmetric. Sampling is used to compensate for this imbalance. To collect a final set of negative samples $S_-=\{(X_i,Y_i)|y_i=0\}$, such that $|S_-|=\gamma|S_+|$, we considered three sampling strategies: random, bootstrapping, and mixture.

Random sampling consists of randomly selecting negative samples according to a uniform distribution. Since the distribution of hard and easy negatives is heavily asymmetric too, most randomly collected samples are easy negatives.
It is well known that hard negatives mining helps boost performance, since hard negatives have the largest influence on the detection accuracy. Bootstrapping accounts for this, by ranking the negative samples according to their objectness scores, and then collecting top $|S_-|$ negatives.  Mixture sampling combines the two, randomly sampling half of $S_-$
and sampling the other half by bootstrapping. In our experiments, mixture sampling has very similar performance to bootstrapping.

To guarantee that each detection layer only detects objects in a
certain range of scales, the training set for the layer
consists of the subset of $S$ that covers the corresponding scale range.
For example, the
samples of smallest scale are used to train the detector of ``det-8''
in Fig. \ref{fig:framework}. It is possible that no positive training samples
are available for a detection layer, resulting in $|S_-|/|S_+|\gg\gamma$.
This can make learning unstable. To address this problem,
the cross-entropy terms of positives and negatives are weighted as follows
\begin{equation}
L_{cls}=\frac{1}{1+\gamma}\frac{1}{|S_+|}\sum_{i\in{S_+}}-\log p_{y_i}(X_i)
+\frac{\gamma}{1+\gamma}\frac{1}{|S_-|}\sum_{i\in{S_-}}-\log p_{0}(X_i).
\end{equation}

\subsection{Implementation Details}
\label{subsec:rpn details}

\subsubsection{Data Augmentation}
In \cite{DBLP:conf/iccv/Girshick15,DBLP:conf/eccv/HeZR014}, it is argued
that multi-scale training is not needed, since deep neural networks are
adept at learning scale invariance. This, however, is not true for datasets
such as Caltech \cite{DBLP:journals/pami/DollarWSP12} and
KITTI \cite{DBLP:conf/cvpr/GeigerLU12}, where object scales can span multiple
octaves. In KITTI, many objects are quite small. Without rescaling, the
cardinalities of the sets $S_+=\{S_+^1,S_+^2,\cdots,S_+^M\}$ are wildly
varying. In general, the set of training examples of largest object size
is very small. To ease this imbalance, the original images are randomly resized
to multiple scales.

\subsubsection{Fine-tuning}
Training the Fast-RCNN \cite{DBLP:conf/iccv/Girshick15} and RPN \cite{DBLP:conf/nips/shaoqing15fasterRcnn} networks requires large amounts of memory and a small mini-batch, due to the large size of the input (i.e. 1000$\times$600). This leads to a very heavy training procedure. In fact, many background regions that are useless for training take substantially amounts of memory. Thus, we randomly crop a small patch (e.g. 448$\times$448) around objects from the whole image. This drastically reduces the memory requirements, enabling four images to fit into the typical GPU memory of 12G.

Learning is initialized with the popular
VGG-Net \cite{DBLP:journals/corr/SimonyanZ14a}. Since bootstrapping
and the multi-task loss can make training unstable in the early iterations,
a two-stage procedure is adopted. The first stage uses random
sampling and a small trade-off coefficient $\lambda$ (e.g. 0.05).
10,000 iterations are run with a learning rate of 0.00005.
The resulting model is used to initialize the second stage,
where random sampling is switched to bootstrapping and
$\lambda=1$. We set $\alpha_i=0.9$ for ``det-8'' and $\alpha_i=1$ for the
other layers. Another 25,000 iterations are run with an initial learning rate of 0.00005, which decays 10 times after every 10,000 iterations. This two-stage learning procedure enables stable multi-task training.

\begin{figure}[!t]
\centering
\centerline{\epsfig{figure=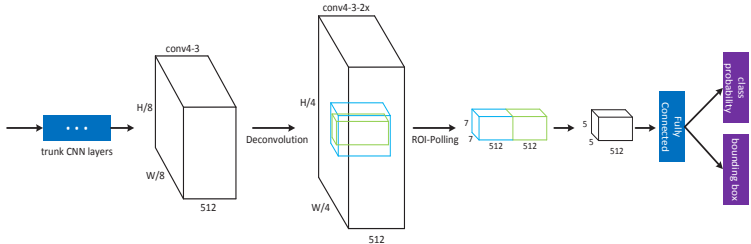,width=10cm,height=3.2cm}}
\caption{Object detection sub-network of the MS-CNN. ``trunk CNN layers" are shared with proposal sub-network. $W$ and $H$ are the
width and height of the input image. The green (blue) cubes represent
object (context) region pooling. ``class probability" and ``bounding box'' are the outputs of the detection sub-network.}
\label{fig:frcnn}
\end{figure}

\section{Object Detection Network}

Although the proposal network could work as a detector itself, it is not
strong, since its sliding windows do not cover objects well. To increase detection accuracy, a detection network is added.
Following \cite{DBLP:conf/iccv/Girshick15}, a ROI pooling layer is first used to extract features of a fixed dimension (e.g. 7$\times$7$\times$512). The
features are then fed to a fully connected layer and output layers, as
shown in Fig. \ref{fig:frcnn}. A deconvolution layer,  described in
Section \ref{subsec:size}, is added to double the resolution of the feature
maps. The multi-task loss of (\ref{equ:multi-task}) is extended to
\begin{equation}
\mathcal{L}(\textbf{W},\textbf{W}_d)
=\sum_{m=1}^{M}\sum_{i\in{S^m}}\alpha_{m}l^{m}(X_i,Y_i|\textbf{W})
+\sum_{i\in{S^{M+1}}}\alpha_{M+1}l^{M+1}(X_i,Y_i|\textbf{W},\textbf{W}_d),
\label{equ:gross loss}
\end{equation}
where $l^{M+1}$ and $S^{M+1}$ are the loss and training samples for the
detection sub-network. $S^{M+1}$ is collected as
in \cite{DBLP:conf/iccv/Girshick15}. As in (\ref{equ:cls and bb}), $l^{M+1}$
combines a cross-entropy loss for classification and a smoothed $L_1$ loss
for bounding box regression. The detection sub-network shares some of the
proposal sub-network parameters $\textbf{W}$ and adds some
parameters $\textbf{W}_d$. The parameters are optimized jointly,
i.e.
$(\textbf{W}^*,\textbf{W}_d^*)=\arg\min\mathcal{L}(\textbf{W},\textbf{W}_d)$.
In the proposed implementation, ROI pooling is applied to the top of
the ``conv4-3'' layer, instead of the ``conv5-3'' layer of
\cite{DBLP:conf/iccv/Girshick15}, since ``conv4-3'' feature maps
performed better in our experiments. One possible explanation is that
``conv4-3'' corresponds to higher resolution and is better suited for
location-aware bounding box regression.

\subsection{CNN Feature Map Approximation}
\label{subsec:size}

Input size has a critical role in CNN-based object detection accuracy.
Simply forwarding object patches, at the original scale, through the CNN
impairs performance (especially for small ones), since the
pre-trained CNN models have a natural scale (e.g. 224$\times$224).
While the R-CNN naturally solves this problem through
warping \cite{DBLP:conf/cvpr/GirshickDDM14}, it is not explicitly
addressed by the Fast-RCNN \cite{DBLP:conf/iccv/Girshick15} or
Faster-RCNN \cite{DBLP:conf/nips/shaoqing15fasterRcnn}. To bridge
the scale gap, these methods simply upsample input images (by $\sim$2 times).
For datasets, such as KITTI \cite{DBLP:conf/cvpr/GeigerLU12}, containing
large amounts of small objects, this has limited effectiveness.
Input upsampling also has three side effects: large memory requirements, slow
training and slow testing. It should be noted that input upsampling does not
enrich the image details. Instead, it is needed because
the higher convolutional layers respond very weakly to small objects.
For example, a 32$\times$32 object is mapped into a 4$\times$4
patch of the ``conv4-3'' layer and a 2$\times$2 patch of the ``conv5-3''
layer. This provides limited information for 7$\times$7
ROI pooling.

To address this problem, we consider an efficient
way to increase the resolution of feature maps. This consists of upsampling
feature maps (instead of the input) using a deconvolution layer,
as shown in Fig. \ref{fig:frcnn}. This strategy is similar to
that of \cite{DBLP:journals/pami/DollarABP14}, shown in Fig.
\ref{fig:multi-scale} (d), where input rescaling is replaced by feature
rescaling. In \cite{DBLP:journals/pami/DollarABP14}, a feature approximator
is learned by least squares. In the CNN world, a better solution is
to use a deconvolution layer, similar to that of
\cite{DBLP:conf/cvpr/LongSD15}. Unlike input upsampling, feature upsampling
does not incur in extra costs for memory and computation. Our
experiments show that the addition of a deconvolution layer significantly
boosts detection performance, especially for small objects.
To the best of our knowledge, this is the first application of deconvolution to jointly improve the speed and accuracy of an object detector.

\subsection{Context Embedding}
\label{subsec:context}

Context has been shown useful for object detection \cite{DBLP:conf/iccv/GidarisK15,DBLP:conf/nips/XiaozhiNIPS15,DBLP:conf/cvpr/BellZBG16} and segmentation \cite{DBLP:conf/cvpr/ZhuUSF15}. Context information has been modeled by a recurrent neural network in \cite{DBLP:conf/cvpr/BellZBG16}
and acquired from multiple regions around the object location in
\cite{DBLP:conf/iccv/GidarisK15,DBLP:conf/nips/XiaozhiNIPS15,DBLP:conf/cvpr/ZhuUSF15}. In this work, we focus on context from multiple
regions.
As shown in Fig. \ref{fig:frcnn}, features from an object (green cube) and
a context (blue cube) region are stacked together immediately after ROI
pooling. The context region is 1.5 times larger than the object region.
An extra convolutional layer without padding is used to reduce
the number of model parameters. It helps compress redundant context and object information, without loss of accuracy, and guarantees that
the number of model parameters is approximately the same.

\subsection{Implementation Details}

Learning is initialized with the model generated by the first learning stage
of the proposal network, described in Section \ref{subsec:rpn details}. The
learning rate is set to 0.0005, and reduced by a factor of 10 times after
every 10,000 iterations. Learning stops after 25,000 iterations. The joint
optimization of (\ref{equ:gross loss}) is solved by back-propagation
throughout the unified network. Bootstrapping is used and $\lambda=1$.
Following \cite{DBLP:conf/iccv/Girshick15}, the parameters of
layers``conv1-1'' to ``conv2-2'' are fixed during learning, for faster
training.


\section{Experimental Evaluation}

The performance of the MS-CNN detector was evaluated on the
KITTI \cite{DBLP:conf/cvpr/GeigerLU12}
and Caltech Pedestrian \cite{DBLP:journals/pami/DollarWSP12} benchmarks.
These were chosen because, unlike
VOC \cite{DBLP:journals/ijcv/EveringhamGWWZ10} and ImageNet
\cite{DBLP:journals/ijcv/RussakovskyDSKS15}, they contain many small objects. Typical image sizes are
1250$\times$375 on KITTI and 640$\times$480 on Caltech.
KITTI contains three object classes: car, pedestrian and cyclist, and three
levels of evaluation: easy, moderate and hard. The ``moderate'' level is the most commonly used. In total, 7,481 images are available for training/validation, and 7,518 for testing.
Since no ground truth is available for the test set, we followed
\cite{DBLP:conf/nips/XiaozhiNIPS15}, splitting the trainval set into training and validation sets. In all ablation experiments, the training set was used for learning and the validation set for evaluation. Following \cite{DBLP:conf/nips/XiaozhiNIPS15}, a model was trained for car detection and another for pedestrian/cyclist detection. One pedestrian model was learned on Caltech. The model configurations for original input size are shown in Table \ref{tab:parameters}. The detector was implemented in C++ within the Caffe toolbox \cite{DBLP:conf/mm/JiaSDKLGGD14}, and source code is available at https://github.com/zhaoweicai/mscnn. All times are reported for implementation on a single CPU core (2.40GHz) of an Intel Xeon E5-2630 server with 64GB of RAM. An NVIDIA Titan GPU was used for CNN computations.

\begin{table}[t]
\centering \scriptsize \setlength{\tabcolsep}{3.0pt}
\vspace{0.1cm} \caption{Parameter configurations of the different models.}
\label{tab:parameters}
\begin{tabular}
{c|c||c|c||c|c||c|c||c||c||c}
\multicolumn{2}{c||}{}
&\multicolumn{2}{|c||}{det-8}
&\multicolumn{2}{|c||}{det-16}
&\multicolumn{2}{|c||}{det-32}
&\multicolumn{1}{|c||}{det-64}
&\multicolumn{1}{|c||}{ROI}
&\multicolumn{1}{|c}{FC}\\\hline\hline
\multirow{2}{*}{car} &filter &5x5 &7x7 &5x5 &7x7 &5x5 &7x7 &5x5 &7x7 &4096\\\cline{2-11}
&anchor &40x40 &56x56 &80x80 &112x112 &160x160 &224x224 &320x320 & &\\\hline
\multirow{2}{*}{ped/cyc} &filter &5x3 &7x5 &5x3 &7x5 &5x3 &7x5 &5x3 &7x5 &2048\\\cline{2-11}
&anchor &40x28 &56x36 &80x56 &112x72 &160x112 &224x144 &320x224 & &\\\hline
\multirow{2}{*}{caltech} &filter &5x3 &7x5 &5x3 &7x5 &5x3 &7x5 &5x3 &8x4 &2048\\\cline{2-11}
&anchor &40x20 &56x28 &80x40 &112x56 &160x80 &224x112 &320x160 & &\\\hline
\end{tabular}
\end{table}

\subsection{Proposal Evaluation}

We start with an evaluation of the proposal network. Following
\cite{DBLP:journals/pami/Hosang2015Pami}, oracle recall is used as
performance metric. For consistency with the KITTI setup, a ground truth
is recalled if its best matched proposal has $IoU$ higher than 70\% for
cars, and 50\% for pedestrians and cyclists.

\begin{table}[t]
\centering \scriptsize \setlength{\tabcolsep}{3.0pt}
\vspace{0.1cm} \caption{Detection recall of the various detection layers on
  KITTI validation set (car), as a function of object hight in pixels.}
\label{tab:layer compraison}
\begin{tabular}
{c||c|c|c|c|c}
\multirow{2}{*}{}
&det-8 &det-16 &det-32 &det-64 &combined \\\hline\hline
25$\leq$height$<$50     &0.9180 &0.3071  &0.0003  &0      &0.9360\\
50$\leq$height$<$100    &0.5934 &0.9660  &0.4252  &0      &0.9814\\
100$\leq$height$<$200   &0.0007 &0.5997  &0.9929  &0.4582 &0.9964\\
height$\geq$200         &0      &0       &0.9583  &0.9792 &0.9583\\\hline
all scales              &0.6486 &0.5654  &0.3149  &0.0863 &0.9611\\\hline
\end{tabular}
\end{table}

\noindent{\bf The roles of individual detection layers\:}
Table \ref{tab:layer compraison} shows the detection accuracy of the
various detection layers as a function of object height in pixels.
As expected, each layer has highest accuracy for the objects
that match its scale. While the individual recall across scales is low,
the combination of all detectors achieves high recall for all object
scales.

\begin{figure*}[!t]
\begin{minipage}[b]{.32\linewidth}
\centering
\centerline{\epsfig{figure=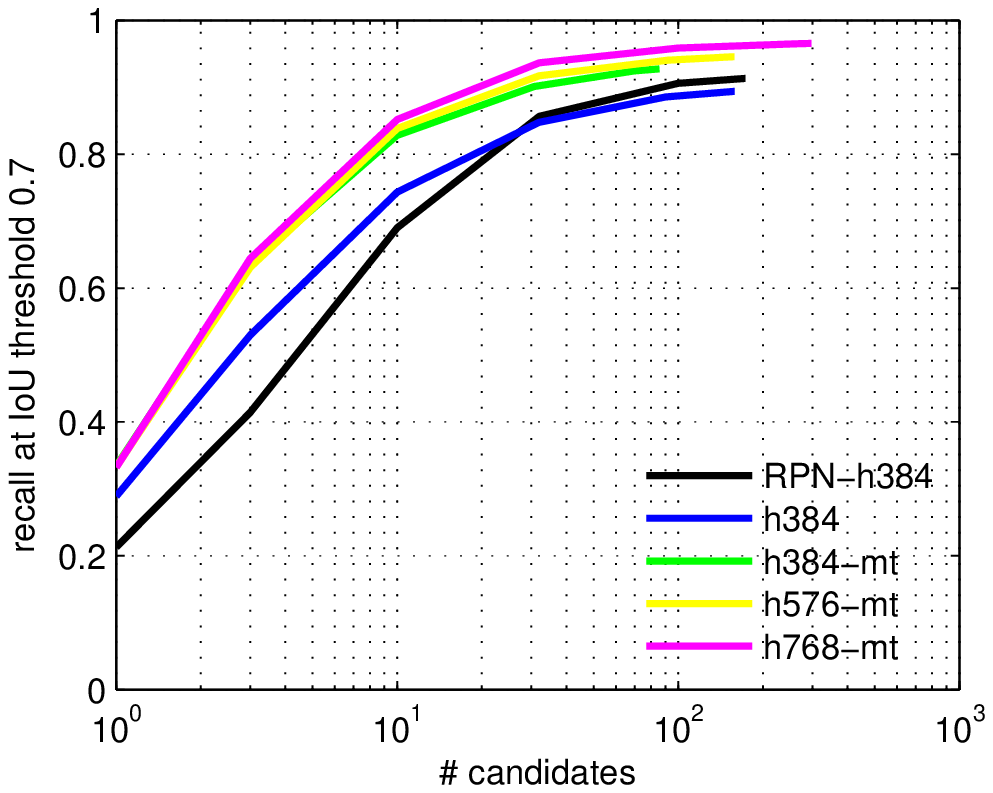,width=3.6cm,height=2.7cm}}{Car}
\end{minipage}
\hfill
\begin{minipage}[b]{.32\linewidth}
\centering
\centerline{\epsfig{figure=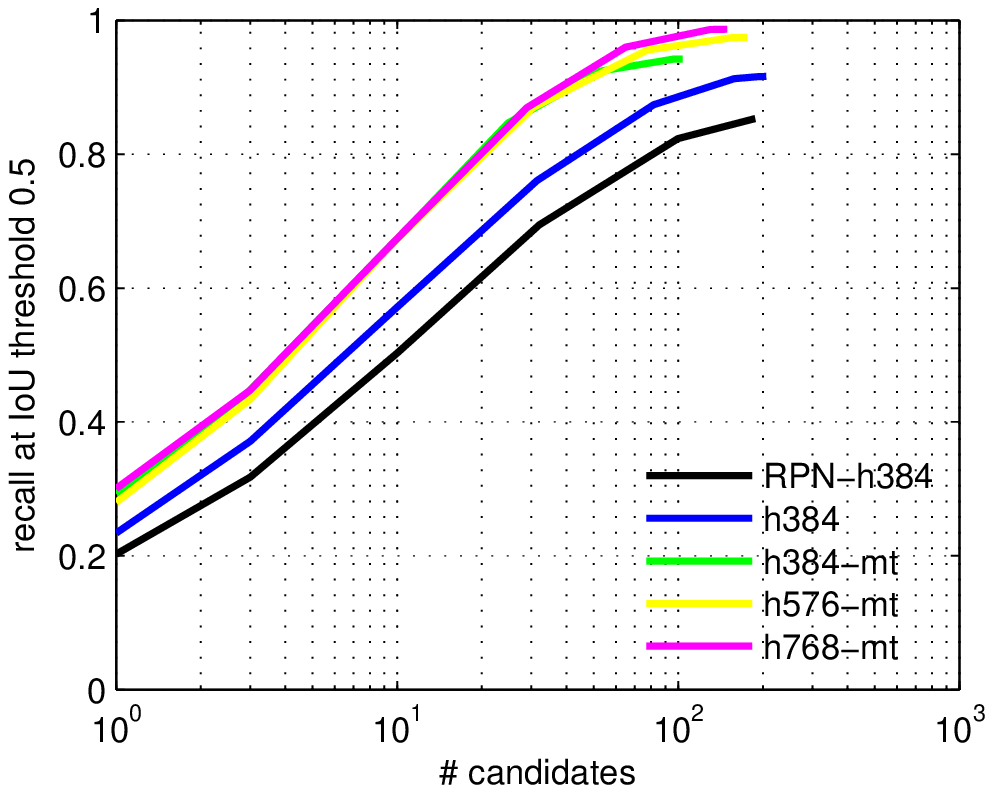,width=3.6cm,height=2.7cm}}{Pedestrian}
\end{minipage}
\hfill
\begin{minipage}[b]{.32\linewidth}
\centering
\centerline{\epsfig{figure=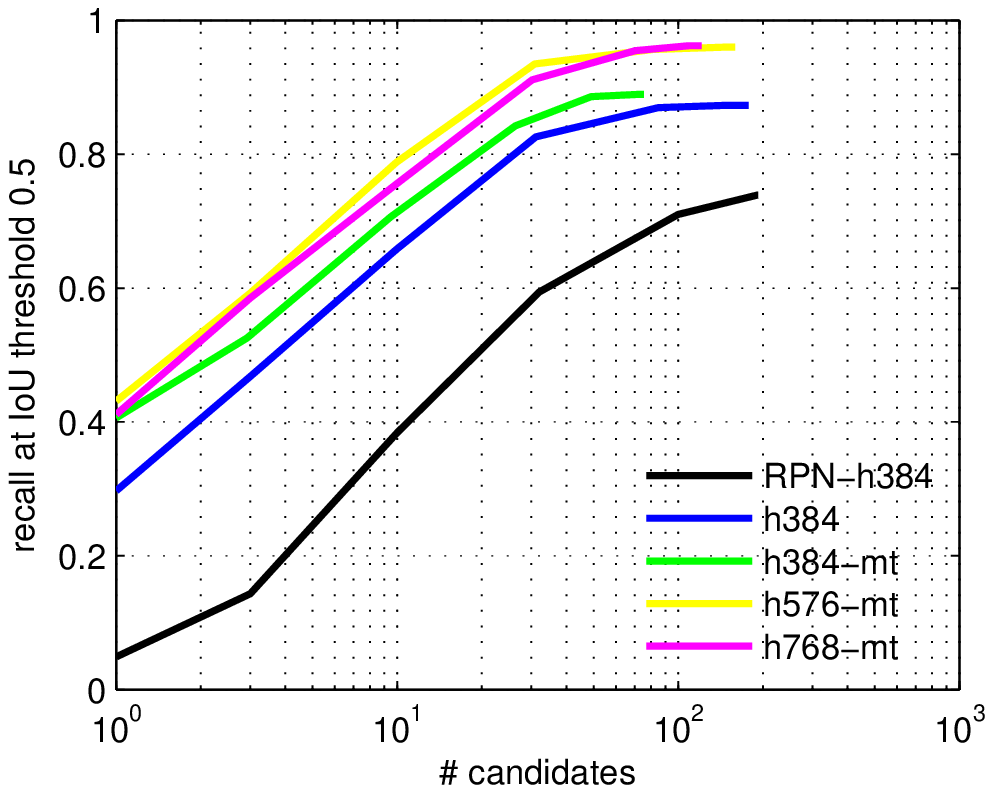,width=3.6cm,height=2.7cm}}{Cyclist}
\end{minipage}
\caption{Proposal recall on the KITTI validation set (moderate).
``hXXX'' refers to input images of height ``XXX''. ``mt'' indicates
multi-task learning of proposal and detection sub-networks.}
\label{fig:mt sz comp}
\end{figure*}

\noindent{\bf The effect of input size\:}
Fig. \ref{fig:mt sz comp} shows that the proposal network is fairly robust
to the size of input images for cars and pedestrians. For cyclist,
performance increases between heights 384 and 576, but there are no gains
beyond this. These results show that the network can achieve good proposal
generation performance without substantial input upsampling.

\noindent{\bf Detection sub-network improves proposal sub-network\:}
\cite{DBLP:conf/iccv/Girshick15} has shown that multi-task learning
can benefit both bounding box regression and classification. On the other
hand \cite{DBLP:conf/nips/shaoqing15fasterRcnn} showed that,
even when features are shared between the two tasks,
object detection does not improve object proposals too much.
Fig. \ref{fig:mt sz comp} shows that, for the MS-CNN, detection
can substantially benefit proposal generation, especially for pedestrians.

\begin{figure*}[!t]
\begin{minipage}[b]{.32\linewidth}
\centering
\centerline{\epsfig{figure=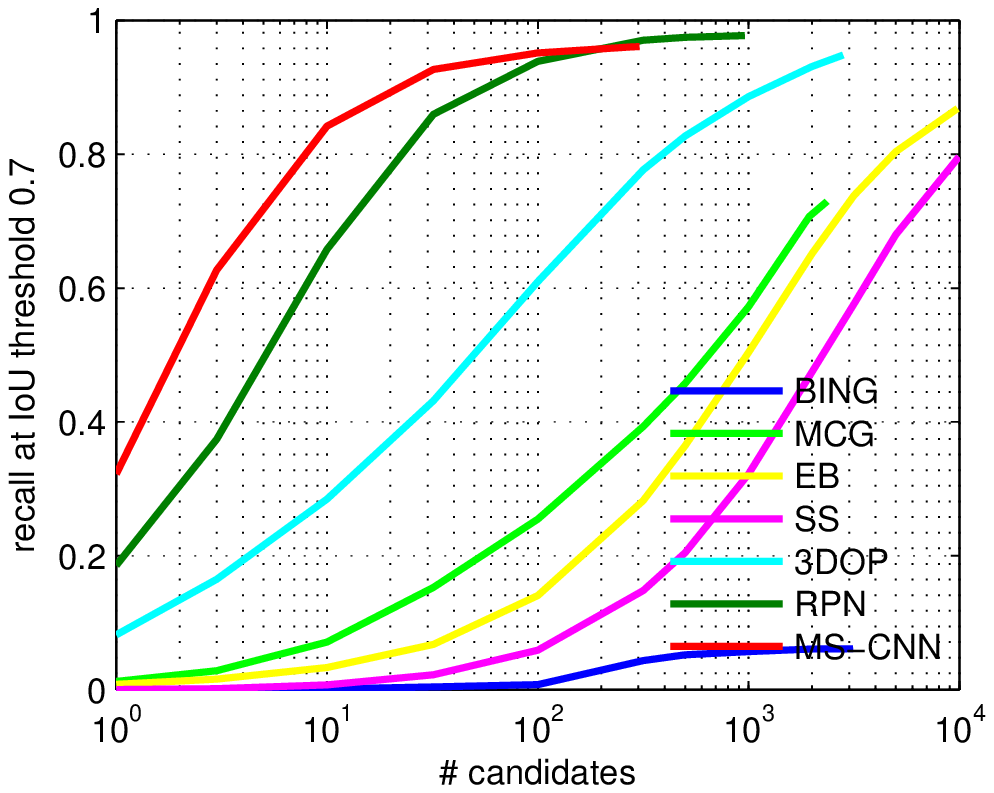,width=3.6cm,height=2.7cm}}
\end{minipage}
\hfill
\begin{minipage}[b]{.32\linewidth}
\centering
\centerline{\epsfig{figure=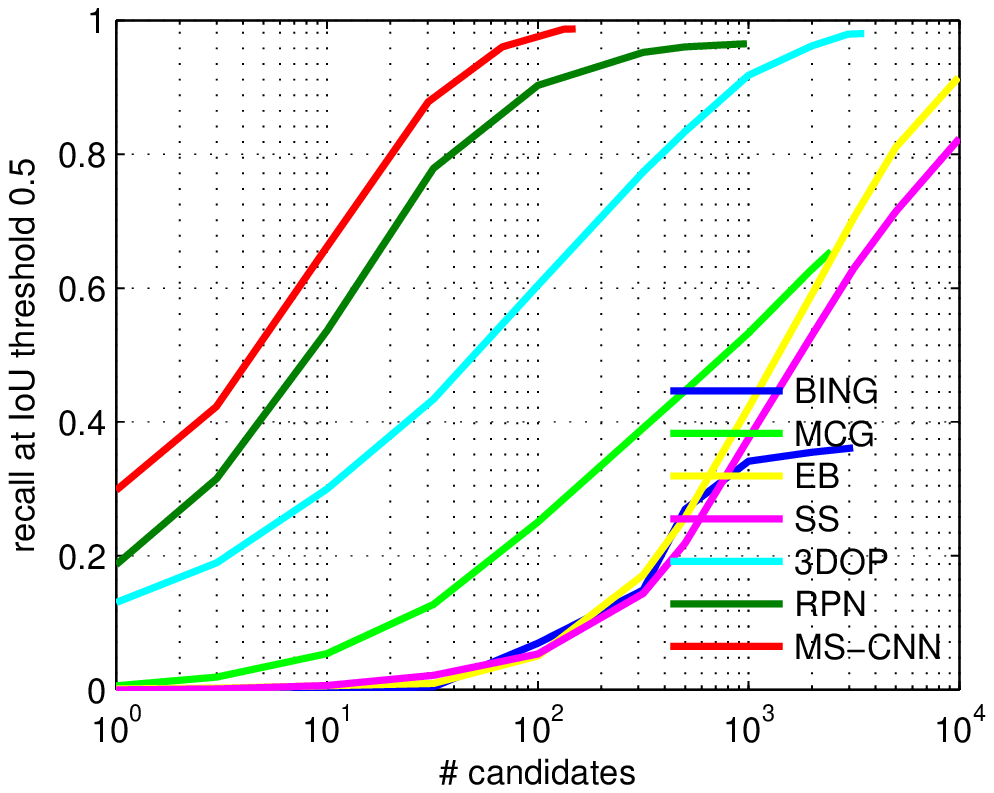,width=3.6cm,height=2.7cm}}
\end{minipage}
\hfill
\begin{minipage}[b]{.32\linewidth}
\centering
\centerline{\epsfig{figure=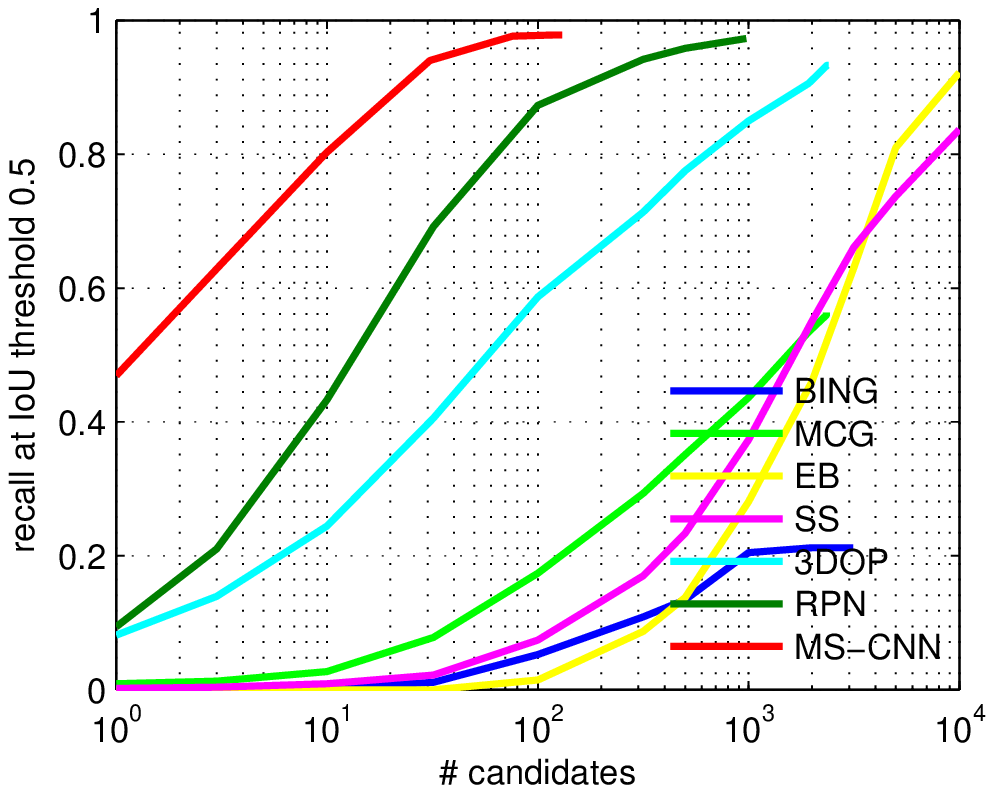,width=3.6cm,height=2.7cm}}
\end{minipage}\\
\hfill
\begin{minipage}[b]{.32\linewidth}
\centering
\centerline{\epsfig{figure=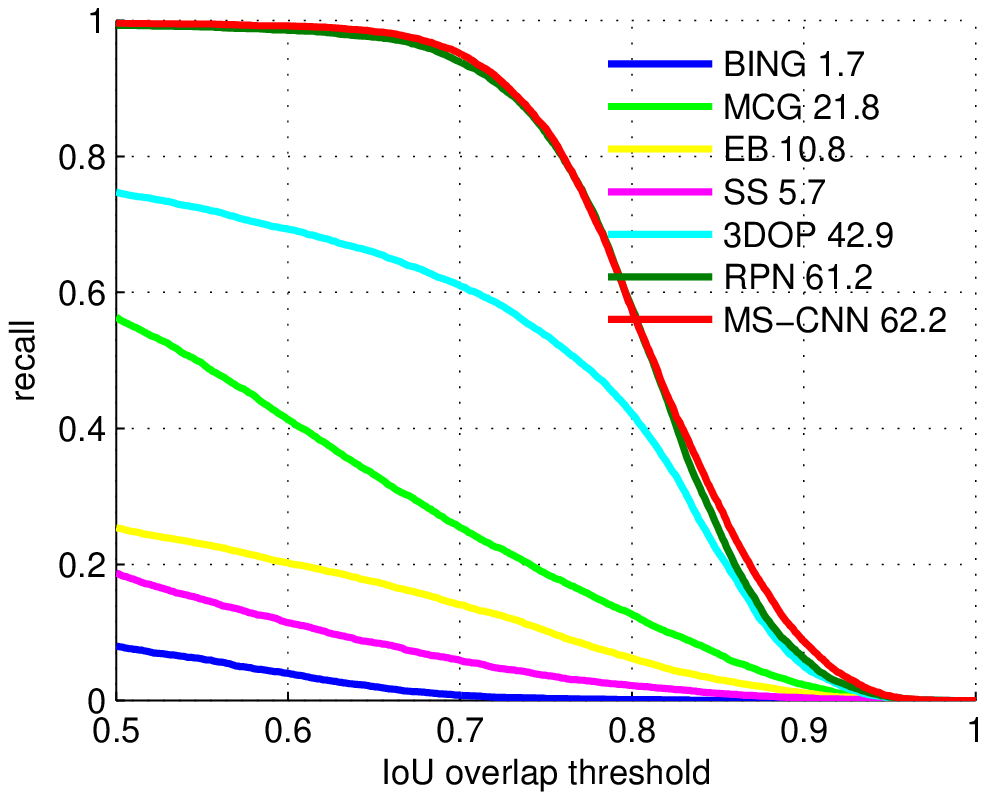,width=3.6cm,height=2.7cm}}{Car}
\end{minipage}
\hfill
\begin{minipage}[b]{.32\linewidth}
\centering
\centerline{\epsfig{figure=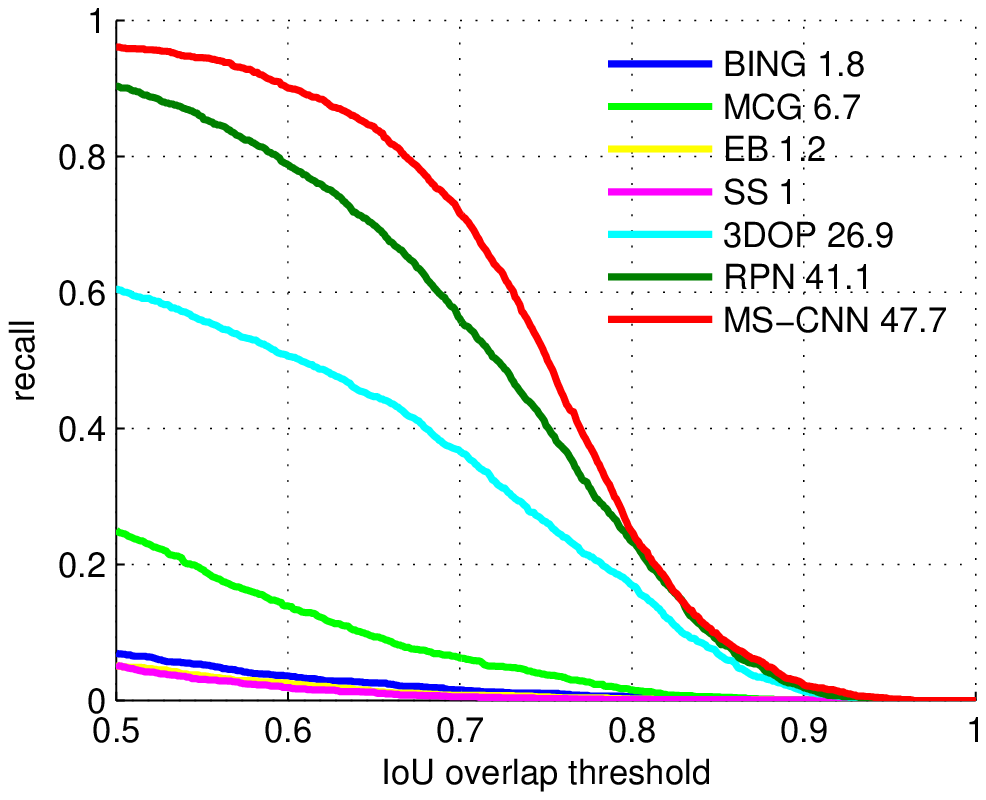,width=3.6cm,height=2.7cm}}{Pedestrian}
\end{minipage}
\hfill
\begin{minipage}[b]{.32\linewidth}
\centering
\centerline{\epsfig{figure=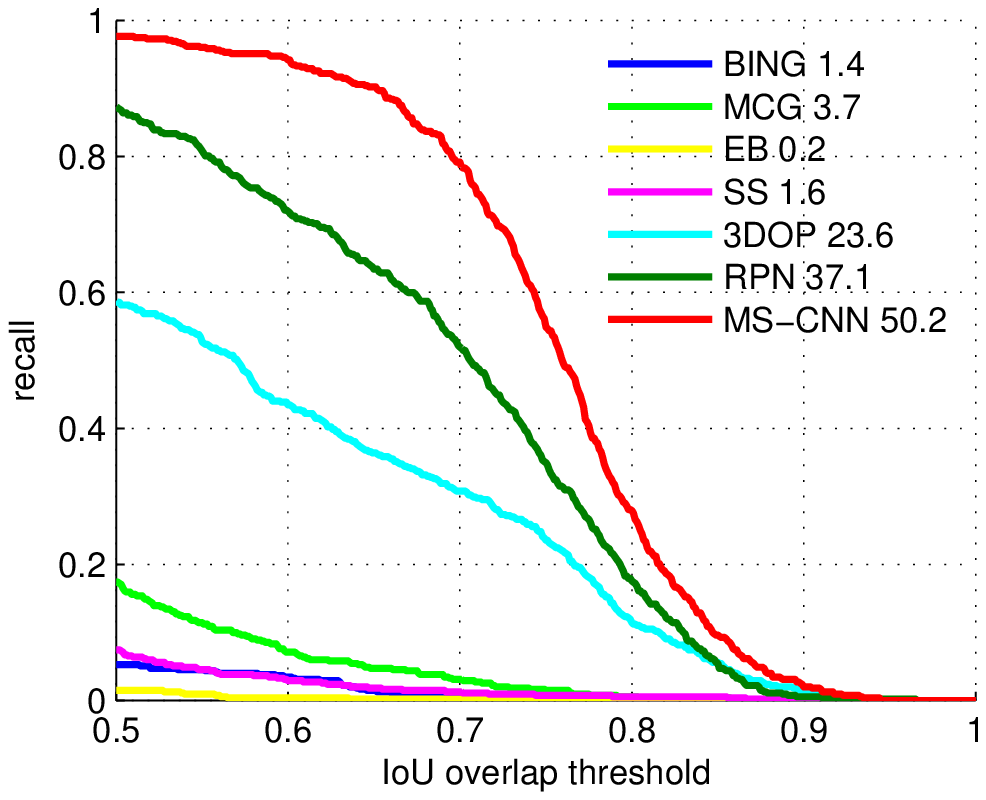,width=3.6cm,height=2.7cm}}{Cyclist}
\end{minipage}
\caption{Proposal performance comparison on KITTI validation set (moderate). The first row is proposal recall curves and the second row is recall v.s. $IoU$ for 100 proposals.}
\label{fig:kitti proposal}
\end{figure*}

\noindent{\bf Comparison with the state-of-the-art\:}
Fig. \ref{fig:kitti proposal} compares the proposal generation network
to BING \cite{DBLP:conf/cvpr/ChengZLT14}, Selective
Search \cite{DBLP:conf/iccv/SandeUGS11},
EdgeBoxes \cite{DBLP:conf/eccv/ZitnickD14},
MCG \cite{DBLP:conf/cvpr/ArbelaezPBMM14},
3DOP \cite{DBLP:conf/nips/XiaozhiNIPS15} and
RPN \cite{DBLP:conf/nips/shaoqing15fasterRcnn}. The top row of the
figure shows that the MS-CNN achieves a recall about 98\%
with only 100 proposals. This should be compared to the $\sim$2,000 proposals
required by 3DOP and the $\sim$10,000 proposals required by EdgeBoxbes.
While it is not surprising that the proposed network outperforms
unsupervised proposal methods, such
as \cite{DBLP:conf/iccv/SandeUGS11,DBLP:conf/eccv/ZitnickD14,DBLP:conf/cvpr/ArbelaezPBMM14}, its large gains over supervised
methods \cite{DBLP:conf/cvpr/ChengZLT14,DBLP:conf/nips/XiaozhiNIPS15},
that can even use 3D information, are significant. The closest
performance is achieved by RPN (input upsampled twice), which has
substantially weaker performance for pedestrians and cyclists.
When the input is not upsampled, RPN misses even more objects, as shown in
Fig. \ref{fig:mt sz comp}. It is worth mentioning that the MS-CNN generates
high quality proposals (high overlap with the ground truth)
without any edge detection or segmentation. This is
evidence for the effectiveness of bounding box regression networks.

\subsection{Object Detection Evaluation}

In this section we evaluate object detection performance.
Since the performance of the cyclist detector has large variance on the
validation set, due to the low number of cyclist occurrences,
only car and pedestrian detection are considered in the ablation experiments.

\begin{table}[t]
\centering \scriptsize \setlength{\tabcolsep}{2.5pt}
\caption{Results on the KITTI validation set. ``hXXX'' indicates an input
of height ``XXX'', ``2x'' deconvolution, ``ctx'' context encoding, and ``c'' dimensionality reduction convolution.  In columns ``Time'' and ``\# params'', entries before the ``/'' are for car model and after for pedestrian/cyclist model.}
\label{tab:model comparison}
\begin{tabular}
{c||c||c||ccc|ccc}\hline\hline
\multirow{2}{*}{Model}
&\multirow{2}{*}{Time}
&\multirow{2}{*}{\# params}
&\multicolumn{3}{|c|}{Cars}
&\multicolumn{3}{|c}{Pedestrians}\\
\cline{4-9}
&\multicolumn{1}{|c||}{} &\multicolumn{1}{|c||}{} & Easy & Mod & Hard & Easy & Mod & Hard \\\hline
h384 &0.11s/0.09s &471M/217M          &90.90 &80.63 &68.94 &73.70 &68.37 &60.72\\
h576 &0.22s/0.19s &471M/217M          &90.42 &88.14 &73.44 &75.35 &70.77 &63.07\\
h768 &0.41s/0.36s &471M/217M          &89.84 &88.88 &75.78 &76.38 &72.26 &64.08\\\hline
h576-random &0.22s/0.19s &471M/217M   &90.94 &87.50 &71.27 &70.69 &65.91 &58.28\\
h576-mixture &0.22s/0.19s &471M/217M  &90.33 &88.12 &72.90 &75.09 &70.49 &62.43\\\hline
h384-2x &0.12s/0.10s &471M/217M       &90.55 &87.93 &71.90 &76.01 &69.53 &61.57\\
h576-2x &0.23s/0.20s &471M/217M       &94.08 &89.12 &75.54 &77.74 &72.49 &64.43\\
h768-2x &0.43s/0.38s &471M/217M       &90.96 &88.83 &75.19 &76.33 &72.71 &64.31\\\hline
h576-ctx &0.24s/0.20s &863M/357M      &92.89 &88.88 &74.34 &76.89 &71.45 &63.50\\
h576-ctx-c &0.22s/0.19s &297M/155M    &90.49 &89.13 &74.85 &76.82 &72.13 &64.14\\\hline
proposal network (h576) &0.19s/0.18s &80M/78M    &82.73 &73.49 &63.22 &64.03 &60.54 &55.07\\\hline
\end{tabular}
\end{table}

\noindent{\bf The effect of input upsampling\:}
Table \ref{tab:model comparison} shows that input upsampling can be a crucial factor for detection. A significant improvement is obtained by upsampling the inputs by 1.5$\sim$2 times, but we saw little gains beyond a factor of 2. This is smaller than the factor of 3.5 required by
\cite{DBLP:conf/nips/XiaozhiNIPS15}. Larger factors lead to (exponentially)
slower detectors and larger memory requirements.

\noindent{\bf Sampling strategy\:}
Table \ref{tab:model comparison} compares sampling strategies: random (``h576-random"), bootstrapping (``h576") and mixture (``h576-mixture"). For car, these three strategies are close to each other. For pedestrian, bootstrapping and mixture are close, but random is much worse. Note that random sampling has many more false positives than the other two.

\noindent{\bf CNN feature approximation\:}
Three methods were attempted for learning the deconvolution layer for feature map approximation: 1) bilinearly interpolated weights; 2) weights initialized by bilinear interpolation and learned with back-propagation; 3) weights initialized with Gaussian noise and learned by back-propagation.
We found the first method to work best, confirming the findings
of \cite{DBLP:conf/cvpr/LongSD15,DBLP:conf/iccv/XieT15}.
As shown in Table \ref{tab:model comparison}, the deconvoltion layer helps
in most cases. The gains are larger for smaller input images, which
tend to have smaller objects. Note that the feature map approximation adds trivial computation and no parameters.

\begin{figure*}[!t]
\begin{minipage}[b]{.32\linewidth}
\centering
\centerline{\epsfig{figure=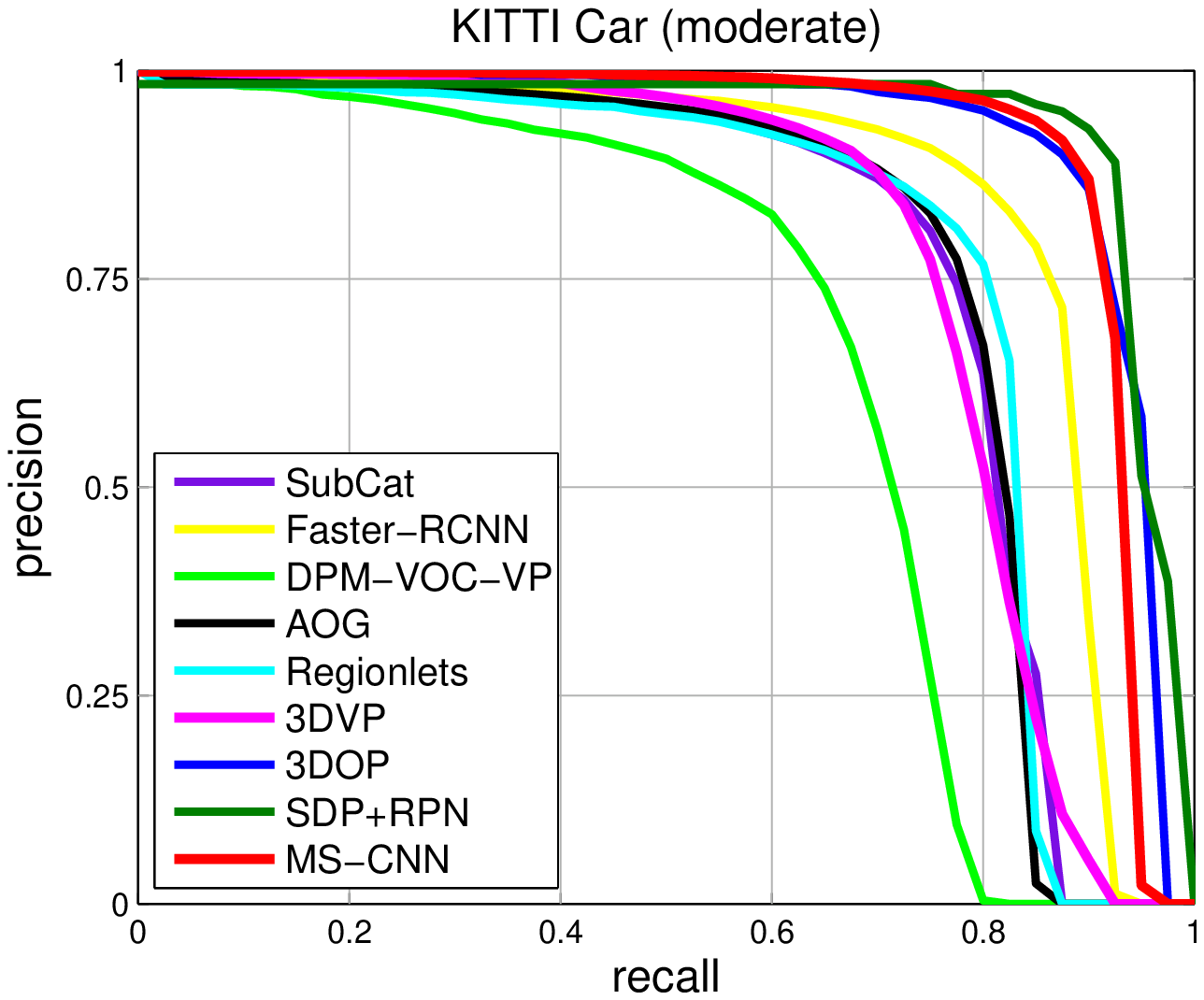,width=4.4cm,height=3.3cm}}
\end{minipage}
\hfill
\begin{minipage}[b]{.32\linewidth}
\centering
\centerline{\epsfig{figure=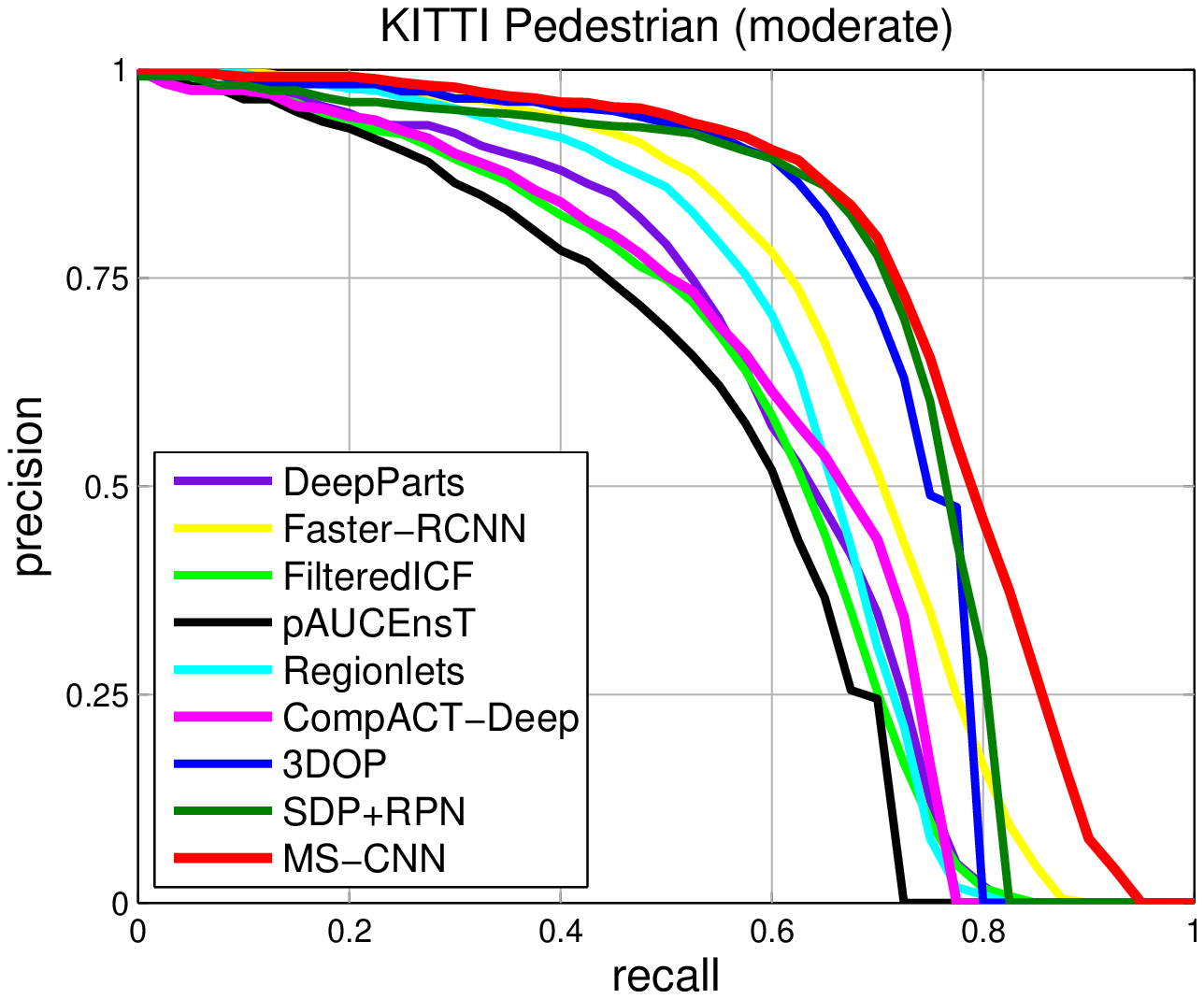,width=4.4cm,height=3.3cm}}
\end{minipage}
\hfill
\begin{minipage}[b]{.32\linewidth}
\centering
\centerline{\epsfig{figure=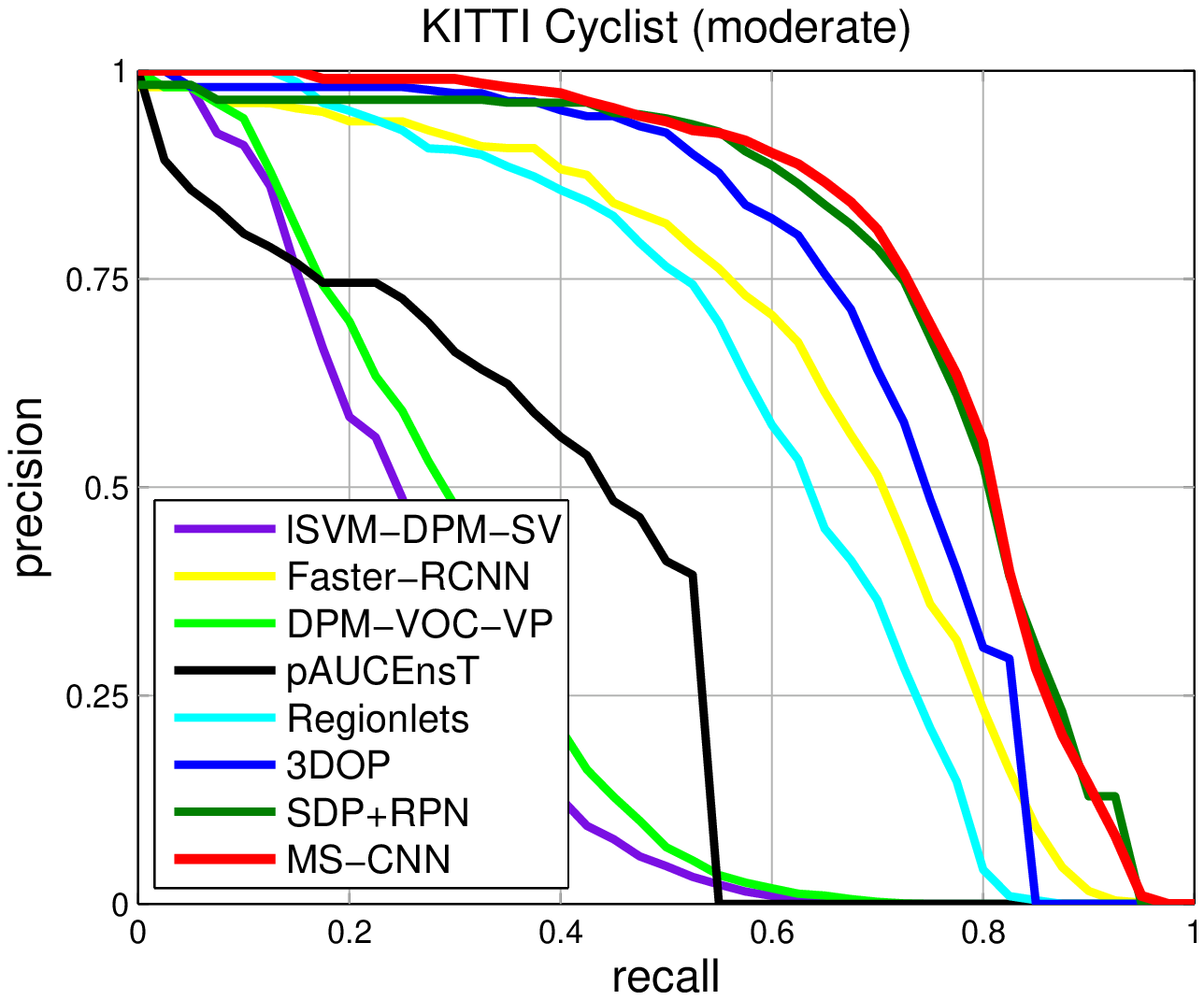,width=4.4cm,height=3.3cm}}
\end{minipage}
\caption{Comparison to the state-of-the-art on KITTI benchmark test set (moderate).}
\label{fig:kitti detection}
\end{figure*}

\begin{table}[t]
\centering \scriptsize \setlength{\tabcolsep}{2pt}
\caption{Results on the KITTI benchmark test set (only published works shown).}
\label{tab:kitti comparison}
\begin{tabular}
{|c||c||ccc||ccc||ccc|}\hline
\multirow{2}{*}{Method}
&\multirow{2}{*}{Time}
&\multicolumn{3}{|c||}{Cars}
&\multicolumn{3}{|c||}{Pedestrians}
&\multicolumn{3}{|c|}{Cyclists}\\
\cline{3-11}
&\multicolumn{1}{|c||}{} & Easy & Mod & Hard & Easy & Mod & Hard & Easy & Mod & Hard \\\hline
LSVM-MDPM-sv \cite{DBLP:conf/nips/GeigerWU11} &10s &68.02 &56.48 &44.18 &47.74 &39.36 &35.95 &35.04 &27.50 &26.21\\
DPM-VOC-VP \cite{DBLP:journals/pami/Pepik0GS15} &8s &74.95 &64.71 &48.76 &59.48 &44.86 &40.37 &42.43 &31.08 &28.23\\
SubCat \cite{DBLP:journals/tits/Ohn-BarT15} &0.7s &84.14 &75.46 &59.71 &54.67 &42.34 &37.95 &- &- &-\\
3DVP \cite{DBLP:conf/cvpr/XiangCLS15} &40s &87.46 &75.77 &65.38 &- &- &- &- &- &-\\
AOG \cite{DBLP:conf/eccv/LiWZ14} &3s &84.80 &75.94 &60.70 &- &- &- &- &- &-\\
Faster-RCNN \cite{DBLP:conf/nips/shaoqing15fasterRcnn} &2s &86.71 &81.84 &71.12 &78.86 &65.90 &61.18 &72.26 &63.35 &55.90\\
CompACT-Deep \cite{DBLP:conf/iccv/CaiSV15}  &1s &- &- &- &70.69 &58.74 &52.71 &- &- &-\\
DeepParts \cite{DBLP:conf/iccv/TianLWT15} &1s &- &- &- &70.49 &58.67 &52.78 &- &- &-\\
FilteredICF \cite{DBLP:conf/cvpr/ZhangBS15} &2s &- &- &- &67.65 &56.75 &51.12 &- &- &-\\
pAUCEnsT \cite{DBLP:journals/corr/PaisitkriangkraiSH14b} &60s &- &- &- &65.26 &54.49 &48.60 &51.62 &38.03 &33.38\\
Regionlets \cite{DBLP:conf/iccv/WangYZL13} &1s &84.75  &76.45 &59.70 &73.14 &61.15 &55.21 &70.41 &58.72 &51.83\\
3DOP \cite{DBLP:conf/nips/XiaozhiNIPS15} &3s  &\textbf{\color{red}93.04} &88.64 &\textbf{\color{red}79.10} &81.78 &67.47 &64.70 &78.39 &68.94 &61.37\\
SDP+RPN \cite{DBLP:conf/cvpr/yang2016sdp} &0.4s &90.14  &88.85 &78.38 &80.09 &70.16 &64.82 &81.37 &73.74 &65.31\\\hline
MS-CNN &0.4s &90.03  &\textbf{\color{red}89.02}  &76.11  &\textbf{\color{red}83.92}  &\textbf{\color{red}73.70}  &\textbf{\color{red}68.31} &\textbf{\color{red}84.06} &\textbf{\color{red}75.46} &\textbf{\color{red}66.07}\\\hline
\end{tabular}
\end{table}

\noindent{\bf Context embedding\:}
Table \ref{tab:model comparison} shows that there is a gain in encoding
context. However, the number of model parameters almost doubles. The
dimensionality reduction convolution layer significantly reduces
this problem, without impairment of accuracy or speed.

\noindent{\bf Object detection by the proposal network\:}
The proposal network can work as a detector, by switching the class-agnostic classification to class-specific. Table \ref{tab:model comparison}
shows that, although not as strong as the unified network, it achieves fairly good results, which are better than those of some detectors on the KITTI leaderboard\footnote{http://www.cvlibs.net/datasets/kitti/}.

\noindent{\bf Comparison to the state-of-the-art\:}
The results of model ``h768-ctx-c'' were submitted to the KITTI leaderboard. A comparison
to previous approaches is given in Table \ref{tab:kitti comparison} and
Fig. \ref{fig:kitti detection}. The MS-CNN set a new record
for the detection of pedestrians and cyclists. The columns ``Pedestrians-Mod'' and ``Cyclists-Mod''
show substantial gains (6 and 7 points respectively) over
3DOP \cite{DBLP:conf/nips/XiaozhiNIPS15}, and much better performance
than the Faster-RCNN \cite{DBLP:conf/nips/shaoqing15fasterRcnn},
Regionlets \cite{DBLP:conf/iccv/WangYZL13}, etc. We also led a nontrivial margin over the very recent SDP+RPN \cite{DBLP:conf/cvpr/yang2016sdp}, which used scale dependent pooling. In terms of speed, the network is fairly fast. For the largest input size, the MS-CNN detector is about 8 times faster than 3DOP. On the original images
(1250$\times$375) detection speed reaches 10 fps.

\begin{figure*}[!t]
\begin{minipage}[b]{.32\linewidth}
\centering
\centerline{\epsfig{figure=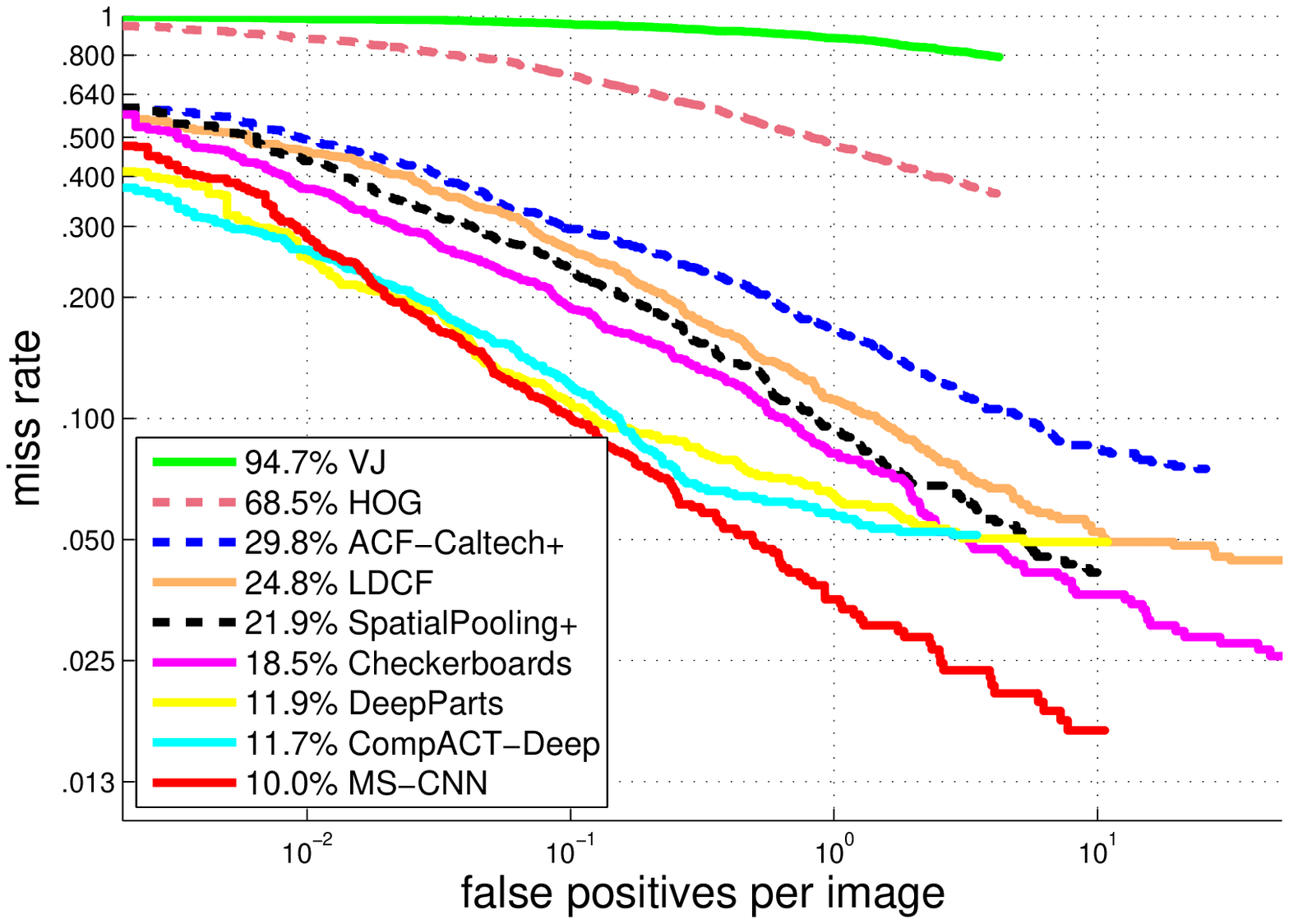,width=4.4cm,height=3.3cm}}{(a) reasonable}
\end{minipage}
\hfill
\begin{minipage}[b]{.32\linewidth}
\centering
\centerline{\epsfig{figure=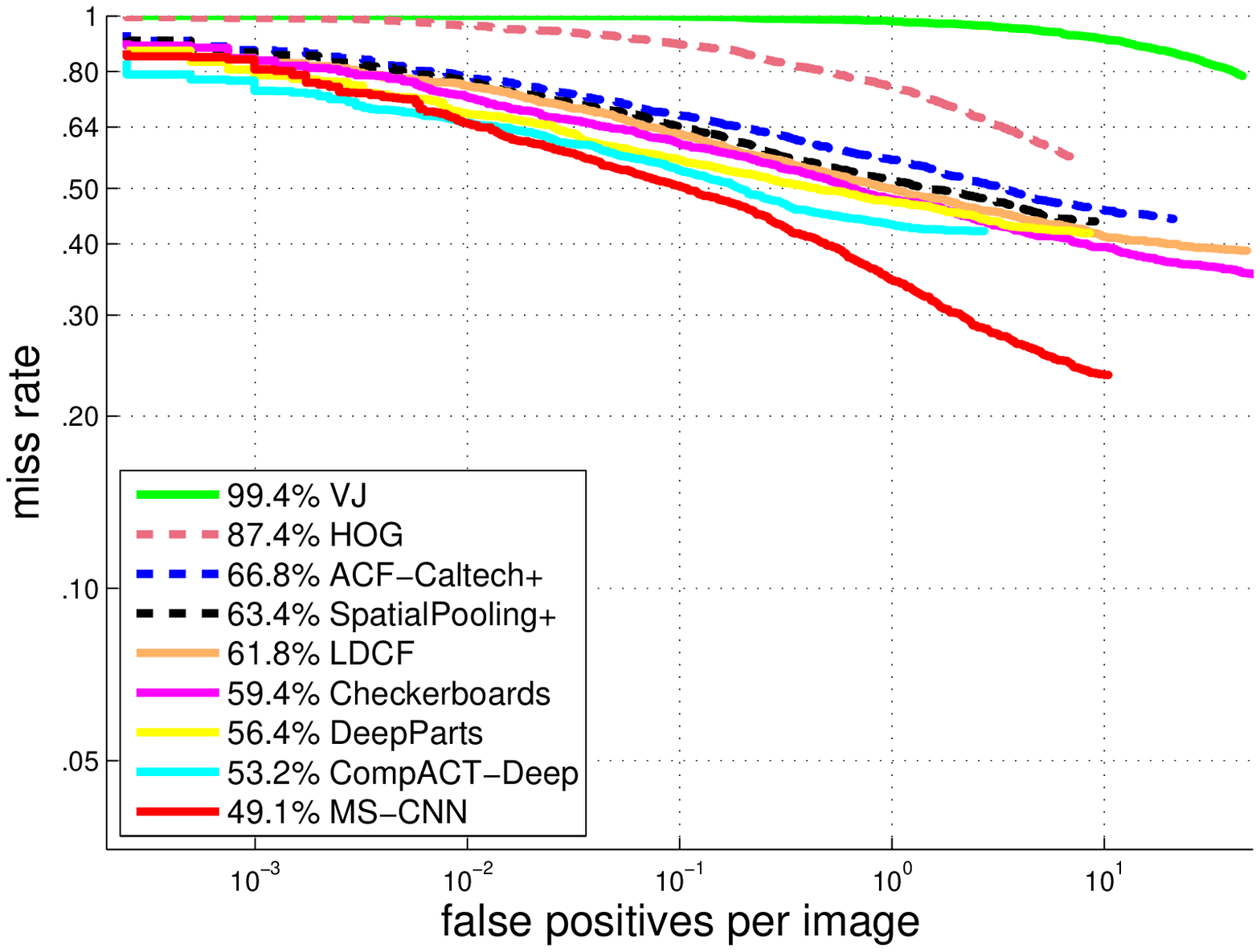,width=4.4cm,height=3.3cm}}{(b) medium}
\end{minipage}
\hfill
\begin{minipage}[b]{.32\linewidth}
\centering
\centerline{\epsfig{figure=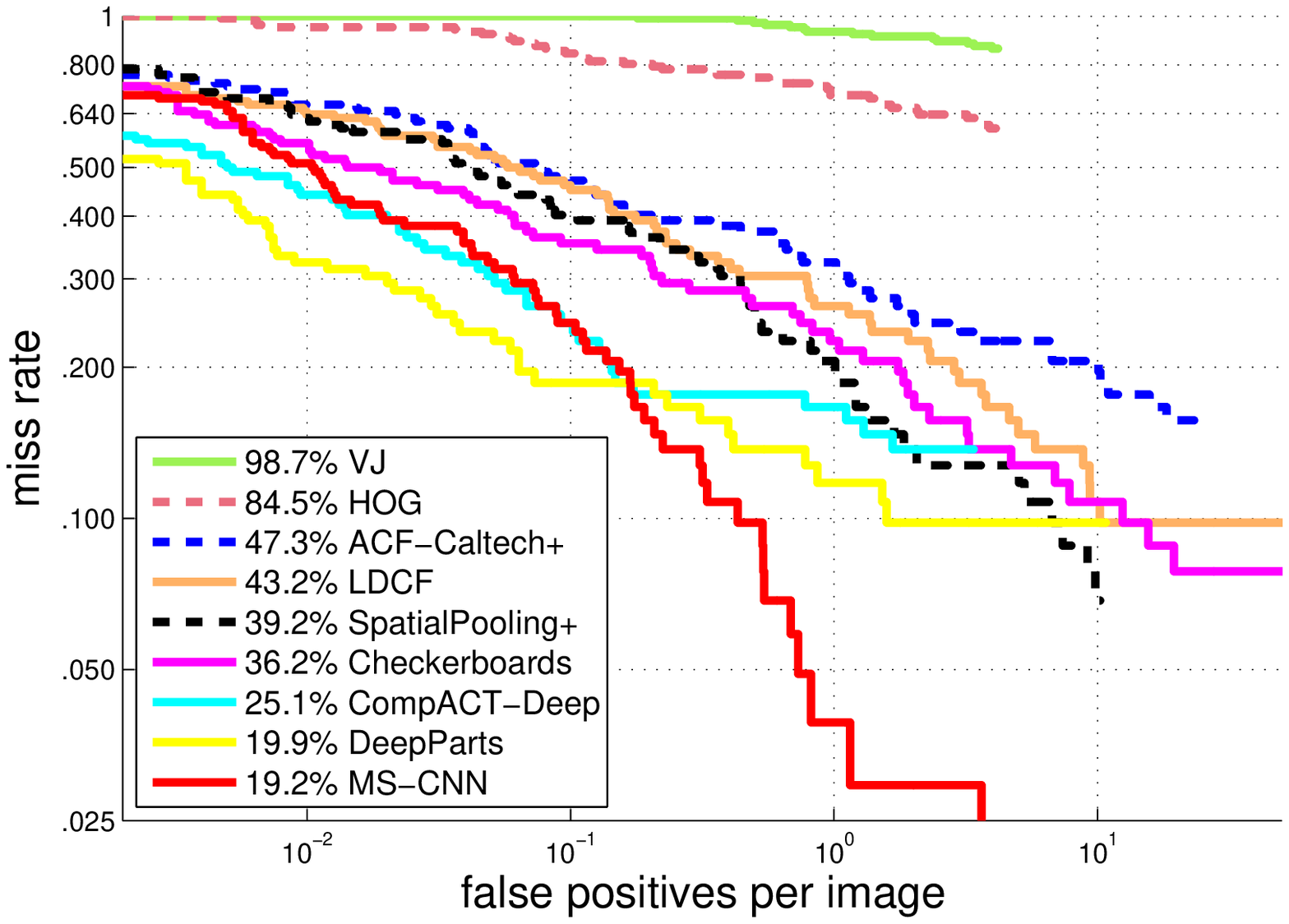,width=4.4cm,height=3.3cm}}{(c) partial occlusion}
\end{minipage}
\caption{Comparison to the state-of-the-art on Caltech.}
\label{fig:caltech}
\end{figure*}

\noindent{\bf Pedestrian detection on Caltech\:}
The MS-CNN detector was also evaluated on the Caltech pedestrian benchmark. The  model ``h720-ctx'' was compared to methods such as
DeepParts \cite{DBLP:conf/iccv/TianLWT15},
CompACT-Deep \cite{DBLP:conf/iccv/CaiSV15},
CheckerBoard \cite{DBLP:conf/cvpr/ZhangBS15},
LDCF \cite{DBLP:conf/nips/NamDH14}, ACF \cite{DBLP:journals/pami/DollarABP14},
and SpatialPooling \cite{DBLP:journals/corr/PaisitkriangkraiSH14b} on
three tasks: reasonable, medium and partial occlusion. As shown in
Fig. \ref{fig:caltech}, the MS-CNN has state-of-the-art performance.
Fig. \ref{fig:caltech} (b) and (c) show that it performs very well for small
and occluded objects, outperforming DeepParts \cite{DBLP:conf/iccv/TianLWT15},
which explicitly addresses occlusion. Moreover, it misses a very small
number of pedestrians, due to the accuracy of the proposal network.
The speed is approximately 8 fps (15 fps) on upsampled
960$\times$720 (original 640$\times$480) Caltech images.

\section{Conclusions}

We have proposed a unified deep convolutional neural network, denoted the
MS-CNN, for fast multi-scale object detection. The detection is preformed at various intermediate network layers, whose receptive fields match
various object scales. This enables the detection of all object scales
by feedforwarding a single input image through the network,
which results in a very fast detector. CNN feature approximation was also
explored, as an alternative to input upsampling. It was shown to
result in significant savings in memory and computation. Overall,
the MS-CNN detector achieves high detection rates at speeds of up to 15 fps.

\subsubsection{Acknowledgement}
This work was partially funded by NSF grant IIS1208522 and a gift from KETI. We also thank NVIDIA for GPU donations through their academic program.

\clearpage

\bibliographystyle{splncs}
\bibliography{egbib}
\end{document}